\documentclass[letterpaper, 10 pt, conference]{ieeeconf}  

\IEEEoverridecommandlockouts                              

\usepackage{amsmath} 
\usepackage{amssymb}  
\usepackage{comment}
\usepackage{kotex}
\usepackage{booktabs}
\usepackage{multirow}
\usepackage{graphicx}
\usepackage{algorithm}
\usepackage{algpseudocode}
\usepackage{stmaryrd}
\usepackage{makecell}
\usepackage{nicefrac}
\usepackage[caption=false]{subfig}
\usepackage{xcolor}
\usepackage{color, colortbl}
\usepackage{soul}
\usepackage{lettrine}
\definecolor{LightCyan}{rgb}{0.88,1,1}

\newcommand{\inc}[1]{{\footnotesize \textcolor{blue}{($+$#1\%)}}}
\newcommand{\dec}[1]{{\footnotesize \textcolor{red}{($-$#1\%)}}}

\newcommand{\tabref}[1]{Tab.~\ref{#1}}
\newcommand{\equref}[1]{Eq.~\ref{#1}}
\newcommand{\figref}[1]{Fig.~\ref{#1}}

\DeclareMathOperator*{\concat}{concat}

\title{\LARGE \bf
Multi-task Learning for Real-time Autonomous Driving \\ Leveraging Task-adaptive Attention Generator
}

\author{Wonhyeok Choi$^{1*}$, Mingyu Shin$^{2*}$, Hyukzae Lee$^{3}$, Jaehoon Cho$^{3}$, Jaehyeon Park$^{3}$, and Sunghoon Im$^{1,2\dag}$
\thanks{$^{1}$W Choi and S Im are with the Department of Electrical Engineering \& Computer Science, DGIST, Daegu, Korea, \texttt{\{smu06117,sunghoonim\}@dgist.ac.kr}}
\thanks{$^{2}$M Shin and S Im are with the Department of Interdisciplinary Studies of Atrificial Intelligence, DGIST, Daegu, Korea, \texttt{\{alsrb4446,sunghoonim\}@dgist.ac.kr}}%
\thanks{$^{3}$H Lee, J Cho, and J Park are with the Autonomous Driving Center, Hyundai Motor Company, Seoul, Korea, \texttt{\{hyukzae.lee, jh.cho, hybrid\}@hyundai.com}}%
\thanks{$^{*}$ Equal-contribution, $^{\dag}$ Corresponding author}
}

\begin{document}

\maketitle
\thispagestyle{empty}
\pagestyle{empty}

\begin{abstract}
Real-time processing is crucial in autonomous driving systems due to the imperative of instantaneous decision-making and rapid response.
In real-world scenarios, autonomous vehicles are continuously tasked with interpreting their surroundings, analyzing intricate sensor data, and making decisions within split seconds to ensure safety through numerous computer vision tasks.
In this paper, we present a new real-time multi-task network adept at three vital autonomous driving tasks: monocular 3D object detection, semantic segmentation, and dense depth estimation.
To counter the challenge of negative transfer — the prevalent issue in multi-task learning — we introduce a task-adaptive attention generator. This generator is designed to automatically discern interrelations across the three tasks and arrange the task-sharing pattern, all while leveraging the efficiency of the hard-parameter sharing approach.
To the best of our knowledge, the proposed model is pioneering in its capability to concurrently handle multiple tasks, notably 3D object detection, while maintaining real-time processing speeds.
Our rigorously optimized network, when tested on the Cityscapes-3D datasets, consistently outperforms various baseline models.
Moreover, an in-depth ablation study substantiates the efficacy of the methodologies integrated into our framework.
\end{abstract}

\vspace{-3pt}
\section{Introduction}

\begin{figure}[t]
  \centering
  \includegraphics[width=\columnwidth]{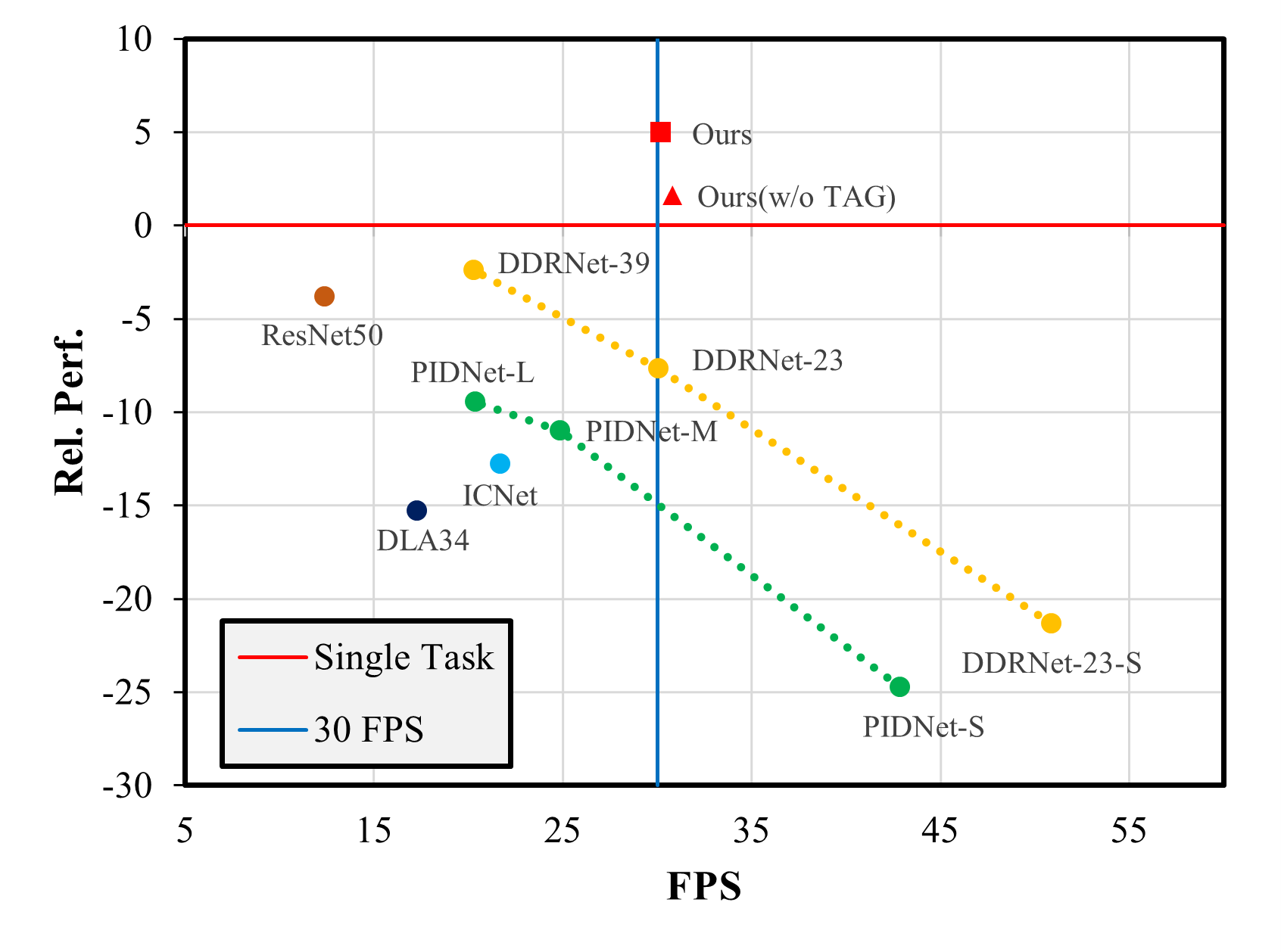}
  \vspace{-25pt}
  \caption{Speed-performance trade-off curve for the Cityscapes-3D validation set (the top-right quadrant is optimal). The Y-axis represents the average relative performance compared to the single-task setup.}
  \label{fig:comparison}
  \vspace{-15pt}
\end{figure}
\lettrine[findent=2pt]{\textbf{W}}{}ith the rise of deep learning technologies, autonomous driving systems have also witnessed remarkable progress in alignment with this development because these perception tasks play a vital role in practical autonomous driving.
Given the inherent nature of practical autonomous driving applications, which require swift and accurate responses in unpredictable environments, the algorithmic speed of deployed models (\textit{e.g.}, latency, inference speed) is the crucial factor.
Hence, training the model individually for each required task (\textit{e.g.}, semantic segmentation or depth estimation) in this system is pragmatically inefficient for these scenarios.

In such scenarios, an applicable learning scheme is Multi-task learning~\cite{ruder2017overview}, which trains multiple tasks within a single model.
This method can improve the generalization ability by consolidating information from various related tasks into a joint feature space.
However, in the context of heterogeneous Multi-Task Learning (MTL), where tasks consisting of different methods are jointly trained, a challenge arises known as the ``negative transfer problem"~\cite{kang2011learning, standley2020tasks}.
This problem can lead to a decrease in overall performance, particularly when dealing with tasks that have low relevance to each other.

Recent works such as soft-parameter sharing~\cite{misra2016cross, ruder2017sluice} and Nas-style MTL~\cite{gao2020mtl, sun2020adashare, maziarz2019flexible, ma2019snr, guo2020learning, fernando2017pathnet} attempt to mitigate this problem by allowing the model to have separate sets of parameters while encouraging some level of parameter similarity or correlation across tasks.
These approaches are particularly useful, as they can mitigate the risk of the negative transfer problem that can occur with hard-parameter sharing.
However, most of these methods use additional network layers or parameters in proportion to the number of tasks, making the time/memory complexity almost equivalent to training multiple single-task models.

\begin{figure*}[t]
  \centering
  \includegraphics[width=\linewidth]{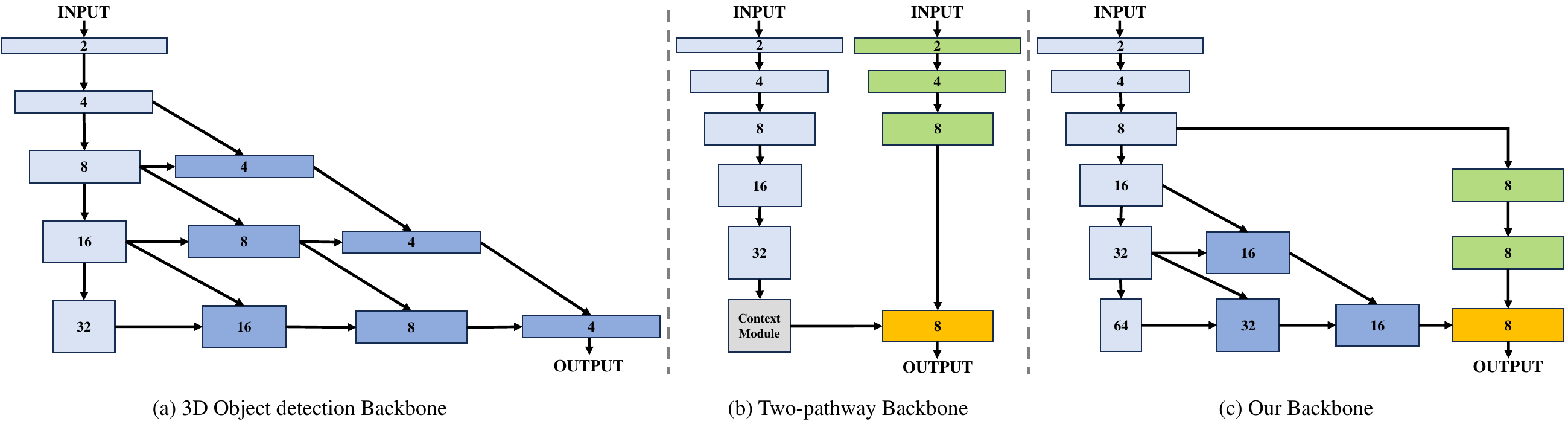}
  \vspace{-20pt}
  \caption{Comparison between existing backbones and our backbone.}
  \label{fig:comparison}
  \vspace{-12pt}
\end{figure*}

To design the real-time multi-task learning framework for autonomous driving, we first adopt the hard-parameter sharing scheme~\cite{jou2016deep, huang2015cross, bilen2016integrated} (a.k.a. shared-bottom), where the layers of the model are explicitly shared across all tasks.
These methods are suffered from negative transfer problems a lot, so we attempt to strike the balance between leveraging shared knowledge across tasks and allowing flexibility for task-adaptive learning while gaining computational efficiency.
We consider two components to achieve this objective: \textit{1. What is the feasible multi-task architecture that can perform important tasks in autonomous driving?}, \textit{2. How to generate task-adaptive features while preserving the computational efficiency of hard-parameter sharing?}.

\noindent\textbf{Multi-task architecture.} Initially, we select the fundamental tasks in autonomous driving as follows: \textit{3D object detection, semantic segmentation, and dense depth estimation.}
Inspired by real-time segmentation architectures~\cite{yu2018bisenet, yu2021bisenet, hong2021deep} and real-time 3D object architecture~\cite{liu2022learning}, we propose a novel backbone that emerges the crucial architectural elements of each task to a hybrid structure.
Our architectural design is based on two-pathway architecture, which is widely used in real-time segmentation frameworks~\cite{yu2018bisenet, hong2021deep} that effectively and efficiently process the per-pixel classification/regression tasks.
To enhance the performance of 3D object detection, where the model needs to estimate the objects with various sizes, we add the aggregation layer to the network that aggregates intermediate features from layers with different receptive fields.
The comparison between existing architectures and ours is illustrated in~\figref{fig:comparison}.

\noindent\textbf{Task-adaptive attention generator.}
According to some soft-parameter sharing MTL~\cite{misra2016cross, ruder2017sluice} and Nas-style MTL~\cite{sun2020adashare, liu2022learning}, conditional computation across tasks such as dynamic neural network~\cite{sun2020adashare} and attention-based method~\cite{liu2019end} can mitigate the negative transfer problem by separating the task-generic and task-specific information.
To enhance the accuracy of our model while preserving the computational efficiency for real-time framework, we propose a Task-adaptive Attention Generator (TAG), which is the attention-based module that generates task-adaptive attention.
Our TAG module can automatically pinpoint the useful feature location for each task along spatial/channel dimensions from the task-generic feature, and it highlights the task-related feature or suppresses the task-unrelated feature, respectively.

In the evaluation results on the Cityscapes 3D, which is the public autonomous driving dataset, our proposed model outperforms the existing baselines with notable margins.
We conduct ablation studies of the proposed architecture and TAG module to validate the effectiveness of the proposed methods.
Also, we report the extensive experiments that show how the TAG module effectively generates task-adaptive features by an attention-based mechanism.
Our contributions are summarized as follows:
\begin{itemize}
    \item To the best of our knowledge, we propose the first real-time autonomous driving framework capable of performing three crucial tasks: 3D object detection, semantic segmentation, and dense depth estimation.
    \item We propose a new architecture and attention-based module for these tasks that alleviate the negative transfer problem while maintaining the model inference cost.
    \item Our framework achieves the best performance and efficiency compared to other baselines consisting of various backbones.
\end{itemize}

\begin{figure*}[t]
  \centering
  \includegraphics[width=\linewidth]{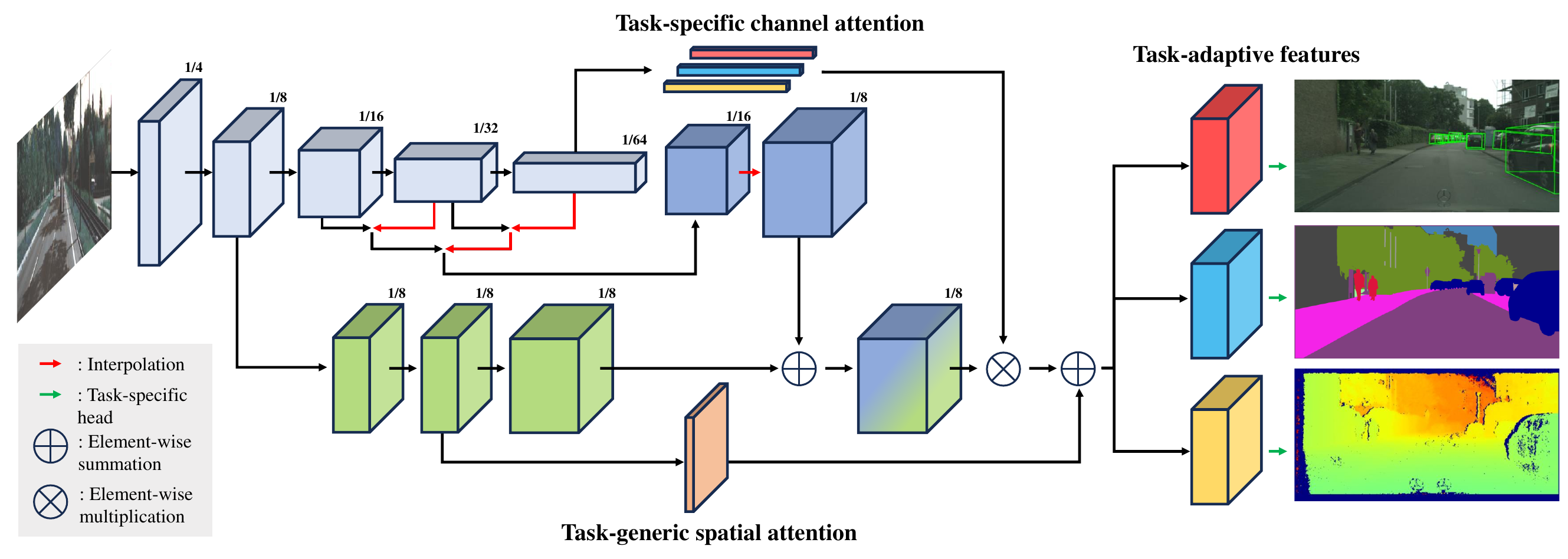}
  \vspace{-22pt}
  \caption{The overall pipeline of the proposed network.}
  \label{fig:pipeline}
  \vspace{-12pt}
\end{figure*}
\section{Related Work}
\subsection{Monocular 3D object detection}
Monocular 3D object detection is broadly categorized into two main approaches.
The first category~\cite{liu2022learning, brazil2019m3d, kumar2022deviant, lu2021geometry, ma2021delving, mousavian20173d, wu2022attention, zhang2021objects} relies solely on RGB images, annotations, and camera calibrations to predict object localizations around a self-driving vehicle.
Most of these methods are based on the foundational work of CenterNet~\cite{zhou2019objects}.
These methods primarily focus on accurately estimating the object depths and encompass various sub-tasks.
The second category compensates for the absence of 3D information in monocular images by utilizing additional data like pre-trained models~\cite{chen2022pseudo, ding2020learning, huang2022monodtr, peng2022did, reading2021categorical, wang2019pseudo} or CAD models~\cite{manhardt2019roi, xiang2015data}.
Some previous works employ pre-trained depth estimators~\cite{ding2020learning, huang2022monodtr, zhang2022monodetr}, either by transforming the monocular settings to LiDAR/Stereo environments~\cite{chen2022pseudo, wang2019pseudo} or dedicating more parameters to the depth estimation task.
Others integrate depth maps from pre-trained depth completion models~\cite{peng2022did} or LiDAR-based detectors~\cite{chen2022pseudo, hong2022cross} for model supervision.
However, while many monocular 3D object detection models exhibit promising outcomes, their slow inference speeds often hinder real-time applications in autonomous driving.
To the best of our knowledge, our model is the first real-time multi-task learning network encompassing monocular 3D object detection.

\subsection{Multi-Task Learning (MTL)}
Hard parameter sharing~\cite{jou2016deep, huang2015cross, bilen2016integrated} is the predominant approach to MTL.
This scheme improves generalization performance and reduces the computational cost of the network by using shared hidden layers between all tasks.
However, a significant downside is its susceptibility to the negative transfer problem~\cite{kang2011learning, standley2020tasks}, which compromises performance when tasks with low relevance are improperly shared.

In contrast, soft-parameter sharing methods~\cite{misra2016cross, ruder2017sluice, liu2019end} address this problem by maintaining separate parameter sets for each task, with looser constraints across tasks.
Nas-style MTL methods have been proposed to adjust shared parameters using a dynamic neural network~\cite{sun2020adashare, maziarz2019flexible, ma2019snr, guo2020learning, fernando2017pathnet} and NAS~\cite{gao2020mtl, raychaudhuri2022controllable, choi2023dynamic}.
Early works use Mixture-of-Experts~\cite{maziarz2019flexible, ma2019snr, fernando2017pathnet} to dynamically adjust the sharing pattern across tasks by selectively choosing the experts from a fixed number of module layers.
Additionally, some works focus on the skip-or-select strategy, opting for task-specific blocks from residual blocks~\cite{sun2020adashare} or designating a shared block for each layer~\cite{guo2020learning, raychaudhuri2022controllable, choi2023dynamic}.
Especially, the study \cite{ye2022taskprompter} integrates prompting to encourage the model to learn the task-specific representation.
While soft-parameter sharing and Nas-style MTL address negative transfer effectively by adapting their parameter-sharing methods, they fall short of offering the computational efficiency inherent in hard parameter sharing.

\vspace{-3pt}
\subsection{Real-time semantic segmentation}
Early works~\cite{orsic2019defense, gamal2018shuffleseg, li2019dfanet} have made advancements in improving the operations in network layers such as depth-wise convolution~\cite{li2019dfanet}, and channel shuffling with group convolution~\cite{gamal2018shuffleseg}.
The study~\cite{orsic2019defense} utilizes both low-resolution and high-resolution images to obtain a high-level semantic understanding.
However, a challenge with these lightweight architectures is the information loss stemming from consistent downsampling.
This hinders the effective retrieval of information during upsampling, resulting in a degradation of the semantic segmentation accuracy.
To mitigate this, more recent works~\cite{yu2018bisenet, yu2021bisenet, hong2021deep} have designed networks with two paths: one low-resolution path for extracting semantic information and another high-resolution path to capture spatial details for boundary delineation.
Similarly, the work \cite{xu2023pidnet} employs a three-branch approach to separately extract details, context, and boundary information. It then utilizes boundary attention to guide the integration of details and context information, resulting in commendable real-time performance.
\section{Method}

In this section, we describe the proposed method, which consists of two main components: the proposed enhanced two-pathway architecture and the Task-adaptive Attention Generator (TAG).
The overall pipeline of our network is illustrated in \figref{fig:pipeline}.

\subsection{Enhanced two-pathway Architecture}
The detailed architecture of the proposed backbone is depicted in \tabref{tab:backbone_config}.
Drawing inspiration from previous works~\cite{yu2018bisenet, hong2021deep}, our architecture consists of two branches: a low-resolution branch for extracting semantic information and a high-resolution branch focused on spatial details.
The high-resolution branch does not contain any downsampling operation to preserve the intricate spatial details. The branching point between the two resolutions is set to layer 3, where the feature map is at 1/8th the resolution of the input image.

Deeper layers within the low-resolution branch extract more semantic and global features. However, this does not confirm that the last layer is the best representation for all tasks. It is worth noting that skip connections are particularly effective for structured tasks like object detection~\cite{hariharan2015hypercolumns, long2015fully, lin2017feature}.
Hence, moving away from the traditional context module, we introduce lightweight hierarchical aggregation layers. These layers pool features from the semantic branch, enhancing the 3D object detection performance.
For the aggregation layers, we redesign the iterative deep aggregation approach~\cite{yu2018deep} to progressively merge multi-stage features and retrieve a high-resolution feature.

\begin{table}[t]
    \caption{Detailed architectures of our backbone for Cityscapes3D.}
    \label{tab:backbone_config}
    \centering
    \resizebox{0.99\linewidth}{!}{
    \begin{tabular}{cccc}
    \toprule
    Stage & Output & \multicolumn{2}{c}{Backbone} \\
    \midrule
    stem & $256 \times 512$ & \multicolumn{2}{c}{$\begin{bmatrix}
        3 \times 3, 64, \text{stride}~ 2 \\
        3 \times 3, 64, \text{stride}~ 2
    \end{bmatrix} \times 1$} \\   
    
    \midrule
    layer1 & $256 \times 512$ & \multicolumn{2}{c}{$\begin{bmatrix}
        3 \times 3, 64 \\
        3 \times 3, 64
    \end{bmatrix} \times 2$} \\  
    
    \midrule
    \multirow{2}{*}{layer2} & \multirow{2}{*}{$128 \times 256$} & \multicolumn{2}{c}{$3 \times 3, 128,$ stride $2$} \\
    \cmidrule{3-4}
    & &
    \multicolumn{2}{c}{[3 $\times$ 3, 128] $\times$ 3} \\

    \midrule
    \multirow{2}{*}{layer3} & \multirow{2}{*}{$64 \times 128, 128 \times 256$} & $3 \times 3, 256,$ stride $2$ & 
    \multirow{2}{*}{$\begin{bmatrix}
        3 \times 3, 128 \\
        3 \times 3, 128
    \end{bmatrix} \times 2$} \\
    \cmidrule{3-3}
    & &
    $[3 \times 3, 256] \times 3$  \\

    \midrule
    \multirow{2}{*}{layer4} & \multirow{2}{*}{$32 \times 64, 128 \times 256$} & $3 \times 3, 512,$ stride $2$ & 
    \multirow{2}{*}{$\begin{bmatrix}
        3 \times 3, 128 \\
        3 \times 3, 128
    \end{bmatrix} \times 2$} \\
    \cmidrule{3-3}
    & &
    $[3 \times 3, 512] \times 3$  \\    

    \midrule
    \multirow{2}{*}{layer5} & \multirow{2}{*}{$16 \times 32, 128 \times 256$} & $3 \times 3, 1024,$ stride $2$ & 
    \multirow{2}{*}{identity} \\
    \cmidrule{3-3}
    & &
    $[3 \times 3, 1024] \times 3$  \\  

    \midrule
    \midrule
    decode layer1 & $32 \times 64, 128 \times 256$ &     
    $\begin{bmatrix}
        3 \times 3, 512 \\
        3 \times 3, 512
    \end{bmatrix} \times 1$ & 
    identity \\   

    \midrule
    decode layer2 & $64 \times 128, 128 \times 256$ &     
    $\begin{bmatrix}
        3 \times 3, 256 \\
        3 \times 3, 256
    \end{bmatrix} \times 2$ & 
    identity \\   
    \midrule
    fusion & $128 \times 256$ & \multicolumn{2}{c}{attention based fusion} \\
    \midrule
    \midrule
    head (MTL) & $1024 \times 2048$ & 
    \multicolumn{2}{c}{$\begin{bmatrix}
        3 \times 3, 256 \\
        1 \times 1, C_t
    \end{bmatrix} \times 2 \times |\mathcal{T}|$ \makecell{\hfill \scriptsize$C_t$: outplane of task $t$\\ \hfill \scriptsize$|\mathcal{T}|$: number of tasks}}\\
    \bottomrule
  \end{tabular}}
  \vspace{-10pt}
\end{table}

As illustrated in Alg.~\ref{alg:cap}, the aggregation layers aggregate the intermediate features of the model moving from shallow layers  to deeper ones, as expressed by the equation:
\begin{align}
\label{eq:aggregation}
f_{ij}(\mathbf{x}, \mathbf{y}) = W_{ij}^{\mathbf{p2}}({\concat}(\mathcal{B}(W_{ij}^{\mathbf{p1}} \mathbf{x}), \mathbf{y})),
\end{align}
where the depth of the aggregation layer $N$ is set to 2. Intermediate features of the semantic branch are characterized by $\mathbf{x}$ and $\mathbf{y}$. For any layer from the node $i$ to $j$, denoted by $f_{ij}$, it is composed of two convolution layers with corresponding weights $W_{ij}^{\mathbf{p1}}, W_{ij}^{\mathbf{p2}}$. 
The process also incorporates a 2$\times$ bilinear interpolation $\mathcal{B}(\cdot)$, followed by a single concatenation operation. 

Lastly, the proposed backbone combines the aggregated feature $\mathbf{x}_{\text{agg}}$ with the output of the high-resolution branch $\mathbf{x}_{\text{detail}}$ to yield a task-generic feature $\mathbf{h}$ as follows:
\begin{align}
\label{eq:head_in}
\mathbf{h} = \mathcal{B}(\mathbf{x}_{\text{agg}}) + g(\mathbf{x}_{\text{detail}}),
\end{align}
where $g$ is the convolution layer that projects the high-resolution feature to match the channel dimension of $\mathbf{x}_{\text{agg}}$.

\subsection{Task-adaptive attention generator (TAG)}
To mitigate the negative transfer problem prevalent in hard-parameter sharing, we propose the TAG module. This module is designed to automatically identify the location of task-adaptive features within the task-generic feature tensor $\mathbf{h}$, which takes the shape of $(B,C,\nicefrac{H}{8},\nicefrac{W}{8})$.
The TAG module operates along the channel and spatial dimensions. Its primary function is to emphasize the task-relevant information while suppressing the unrelated details.
For channel attention, TAG receives the last feature of the semantic branch $\mathbf{x}_{\text{semantic}}$ to capture the model's most semantic and global feature.
On the other hand, the TAG module derives task-generic spatial attention from the output of the high-resolution branch $\mathbf{x}_{\text{detail}}$ as follows:
\begin{align}
\begin{split}
\label{eq:attn}
\alpha_{t} = \sigma(S_{ch}^{t}(\mathbf{x}_{\text{semantic}})), ~
\beta = \sigma(S_{sp}(\mathbf{x}_{\text{detail}})),
\end{split}
\end{align}
where $t \in \{0, 1, 2\}$ is the task index. The operation $\sigma$ is the sigmoid function. $S_{ch}^{t}$ is the channel attention layer consisting of two MLP layers following global average pooling. Meanwhile, $S_{sp}$ is the task-generic spatial attention layer that includes single dilated convolution.
Through \equref{eq:attn}, we obtain both task-specific channel attention and task-generic spatial attention, represented as $\alpha_{t}$ and $\beta$, respectively.

The TAG module constructs the task-adaptive features $\mathbf{h}_{t}$ by performing channel-wise multiplication between the task-specific channel attention map $\alpha_{t}$ and the task-generic feature $\mathbf{h}$, followed by an addition of the spatial attention $\beta$ as follows:
\begin{align}
\label{eq:tag_out}
\mathbf{h}_{t} = \alpha_{t} \cdot \mathbf{h} + \beta.
\end{align}

\begin{algorithm}[t]
\caption{Forward procedure of aggregation layer}
\label{alg:cap}
\begin{algorithmic}
\Require $L = [\mathbf{x}_{1},...,\mathbf{x}_{N}]$ \Comment{{\footnotesize list contains $N$ Multi-scale features}}
\Ensure $\mathbf{x}_{\text{agg}}$ \Comment{{\footnotesize aggregated feature}}
\State idx$_1$, idx$_2$ $ \gets 1,0$ \Comment{{\footnotesize layer/scale index}}
\While{idx$_1$ $\neq N$}
\State idx$_2$ $\gets$ idx$_1$
\While{idx$_2$ $\neq 0$}
\State $L[\text{idx$_2$}-1] \gets f_{\text{idx$_1$,idx$_2$}}(L[\text{idx$_2$}-1], L[\text{idx$_2$}])$
\State idx$_2$ $\gets$ idx$_2-1$ \Comment{{\footnotesize refer to \equref{eq:aggregation}}}
\EndWhile
\State idx$_1$ $\gets$ idx$_1+1$
\EndWhile
\State $\mathbf{x}_{\text{agg}} \gets L[0]$
\end{algorithmic}
\end{algorithm}

\vspace{-5pt}
\subsection{Multi-head Structure}
3D object detection frameworks typically consist of multiple sub-tasks, including classification, center localization, and heading direction estimation.
In our approach, we integrate heads for various roles: sub-tasks related to 3D object detection, semantic segmentation, and dense depth estimation to maintain a compact model structure. All these heads employ shallow CRC layers, which are composed of 3$\times$3Conv-ReLU-1$\times$1Conv.
For the sub-task associated with 3D detection, we basically adopt the approach of \cite{ma2021delving}, but with an added functionality to deduce the pitch and roll, facilitating the inference of the 3-DoF heading direction.

\subsection{Loss function}
For the 3D detection loss, our model is largely aligned with \cite{ma2021delving}. However, we introduce modifications for the pitch and roll tasks. Specifically, the heading loss for these tasks is informed by the method proposed in~\cite{wang2021fcos3d}.
As for the semantic segmentation and dense depth estimation tasks, we employ cross-entropy and smooth L1 loss, respectively.
In setting the loss weights for our three tasks, we follow the scheme presented in \cite{ye2022taskprompter}. Here, the weights $\lambda_{det}, \lambda_{seg}, \lambda_{dep}$ are set to $1$, $100$, and $1$, respectively.
Our final loss term is formulated as follows:
\begin{align}
\label{eq:loss}
\mathcal{L}_{task} = \lambda_{det} \mathcal{L}_{det} + \lambda_{seg} \mathcal{L}_{seg} + \lambda_{dep} \mathcal{L}_{dep}.
\end{align}

\vspace{-10pt}
\section{Experiments}
\sethlcolor{LightCyan}
\begin{table*}[t]
    \caption{Evaluation on Cityscapes-3D validation set (\textbf{best}, \underline{second-best}, \hl{$\geq$30FPS},  \textcolor{blue}{positive}/\textcolor{red}{negative} performance deviation from STL).}
    \label{tab:Cityscapes-3D}
    \centering
    \begin{tabular}{cclllrccc}
    \toprule
    \multirow{2}{*}{Method} & \multirow{2}{*}{Backbone} & 3D Det. & Semantic seg. & Dense depth & \multirow{2}{*}{$\Delta_{\mathcal{T}}$ (\%) $\uparrow$} & \multirow{2}{*}{FPS $\uparrow$} & \multirow{2}{*}{GFLOPs $\downarrow$} & \multirow{2}{*}{Params (M) $\downarrow$} \\
    \cmidrule{3-5}
    & & DS $\uparrow$ & mIoU $\uparrow$ & RMSE $\downarrow$ & & & & \\
    \midrule
    \multirow{3}{*}{STL} & \multirow{3}{*}{DLA34} & 54.55 & - & - & 0.00 & 17.4 & 246 & 20.1 \\
    & & - & 66.38 & - & 0.00 & 17.8 & 222 & 19.9 \\
    & & - & - & 5.50 & 0.00 & 17.8 & 223 & 19.9 \\
    \midrule
    \multirow{9}{*}{MTL} & DLA34 & 38.05 \dec{30.25} & 63.79 \hspace{1.4pt} \dec{3.90} & 6.15 \dec{11.82} & \textcolor{red}{$-$15.32} & 17.3 & 285 & 20.4 \\
        & ResNet50               & 51.92 \hspace{1.4pt} \dec{4.82} & 66.40 \hspace{1.4pt} \inc{0.03} & 5.87 \hspace{1.4pt} \dec{6.73} & \textcolor{red}{$-$3.84} & 12.4 & 373 & 44.5 \\
        & ICNet                  & 39.67 \dec{27.28} & 61.18 \hspace{1.4pt} \dec{7.83} & 5.68 \hspace{1.4pt} \dec{3.27} & \textcolor{red}{$-$12.79} & 21.7 & 226 & 44.4 \\
        & DDRNet-23-slim         & 32.36 \dec{40.68} & 60.06 \hspace{1.4pt} \dec{9.52} & 6.26 \dec{13.82} & \textcolor{red}{$-$21.34} & \cellcolor{LightCyan}50.9 & 75 & 7.5 \\
        & DDRNet-23              & 49.19 \hspace{1.4pt} \dec{9.83} & 62.80 \hspace{1.4pt} \dec{5.39} & 5.93 \hspace{1.4pt} \dec{7.82} & \textcolor{red}{$-$7.68} & \cellcolor{LightCyan}30.1 & 216 & 23.9 \\
        & DDRNet-39              & 50.83 \hspace{1.4pt} \dec{6.82} & \textbf{67.17} \hspace{1.4pt} \inc{1.19} & 5.59 \hspace{1.4pt} \dec{1.64} & \textcolor{red}{$-$2.42} & 20.3 & 345 & 35.5 \\
        & PIDNet-S               & 40.12 \dec{26.45} & 50.57 \dec{23.82} & 6.82 \dec{24.00} & \textcolor{red}{$-$24.76} & \cellcolor{LightCyan}42.9 & 84 & 9.4 \\
        & PIDNet-M               & 50.78 \hspace{1.4pt} \dec{6.91} & 58.31 \dec{12.16} & 6.27 \dec{14.00} & \textcolor{red}{$-$11.02} & 24.9 & 241 & 32.2 \\
        & PIDNet-L               & 53.48 \hspace{1.4pt} \dec{1.96} & 58.81 \dec{11.40} & 6.33 \dec{15.09} & \textcolor{red}{$-$9.48} & 20.4 & 339 & 40.3 \\
    \midrule
    \multirow{2}{*}{MTL} & Ours (w/o TAG) & \underline{57.46} \hspace{1.4pt} \inc{5.33} & 65.90 \hspace{1.4pt} \dec{0.72} & \underline{5.49} \hspace{1.4pt} \inc{0.18} & \textcolor{blue}{\underline{$+$1.60}} & \cellcolor{LightCyan}30.8 & 218 & 32.8 \\
                         & Ours           & \textbf{61.21} \inc{12.21} & \underline{66.50} \hspace{1.4pt} \inc{0.18} & \textbf{5.36} \hspace{1.4pt} \inc{2.55} & \textcolor{blue}{\textbf{$+$4.98}} & \cellcolor{LightCyan}30.2 & 219 & 33.9 \\
    \bottomrule
  \end{tabular}
  \vspace{-5pt}
\end{table*}

\subsection{Dataset}
\subsubsection{Cityscapes-3D}
Cityscapes-3D~\cite{gahlert2020cityscapes} is an extended version of the Cityscapes dataset~\cite{cordts2016cityscapes} by incorporating additional 3D bounding box annotations.
It retains the original Cityscapes configuration with 2,975 training images and 500 validation images, all finely annotated at a resolution of 1024$\times$2048.

In contrast to previous 3D object detection datasets, such as nuScenes~\cite{caesar2020nuscenes}, Waymo~\cite{sun2020scalability}, and KITTI~\cite{Geiger2012CVPR}, Cityscapes-3D has been designed specifically for the purpose of monocular 3D object detection.
This distinction arises because Cityscapes-3D relies on ground-truth depth labels from stereo cameras, rather than LiDAR. This often introduces disparities between images and depth maps.
Moreover, Cityscapes-3D provides labels for both pitch and roll angles of vehicles, whereas many previous datasets primarily concentrate on yaw. 
Recognizing these angles is crucial in instances where roads are not flat, a common occurrence.


\subsection{Baselines}
We compare our model to various backbones with multi-head structures to validate the effectiveness of the proposed architecture and TAG module.
We select five well-known baselines as comparison sets: ResNet50~\cite{he2016deep}, DLA34~\cite{yu2018deep}, ICNet~\cite{zhao2018icnet}, DDRNet~\cite{hong2021deep}, and PIDNet~\cite{xu2023pidnet}.


\subsection{Evaluation Metric}
For the task of 3D object detection, we employ the Detection Score, which is the official evaluation index provided by Cityscapes-3D.
In the context of semantic segmentation, the mean Intersection over Union (mIoU) serves as our metric, while for dense depth estimation, we use the Root Mean Square Error (RMSE) for evaluation.
To understand the in/decremental shift in overall task performance from a single-task setup, we adopt the averaged relative performance $\Delta_{\mathcal{T}}$ in~\cite{sun2020adashare} as follows:
\begin{align}
\begin{split}
\label{eq:rel_performance1}
\Delta_{\mathcal{T}} = {\frac{100}{|\mathcal{T}|}} \sum_{t \in \mathcal{T}}(-1)^{l_t}{\frac{(\mathcal{M}_{t} - \mathcal{M}_{t}^{single})}{\mathcal{M}_{t}^{single}}}, 
\end{split}
\end{align}
where $\mathcal{M}_{t}$ and $\mathcal{M}_{t}^{single}$ are the metric of task $t$ from each method and the single task baseline, respectively.
The constant $l_t$ is 1 if a lower value represents better for the metric $\mathcal{M}_{t}$ and 0 otherwise.

\subsection{Implementation details}
Like all previous baselines~\cite{hong2021deep, xu2023pidnet, yu2018deep, he2016deep}, we initialize our model by pretraining on ImageNet~\cite{russakovsky2015imagenet}.
All models undergo training with an input resolution set at 224$\times$224, a batch size of 256, spanning a total of 100 epochs.
We commence with a learning rate of 0.1, and reduce it by a factor of 10 during the 30th, 60th, and 90th epochs.
All networks are trained using SGD with a weight decay of $1$$\times$$ 10^{-4}$ and Nesterov momentum set to 0.9.

For Cityscapes-3D training, the same multi-head structure and configuration are maintained across all baselines to ensure a fair comparison, differing only in the backbone (\textit{e.g.}, $\mathcal{L}_{task}$, loss scale, and learning rate).
Training parameters include an input resolution at the original 1024$\times$2048 dimensions, a batch size of 2, and no image augmentation.
For all our baseline models, the Adam optimizer is employed with an initial learning rate of $2$$\times$$10^{-5}$.
Moreover, we employ a polynomial learning rate policy, with a power of 0.9, for a progressive reduction in the learning rate.



\begin{figure*}[t]
  \centering
  \subfloat[Visualization of task-generic/adaptive features.\label{1a}]{
       \includegraphics[width=0.55\linewidth]{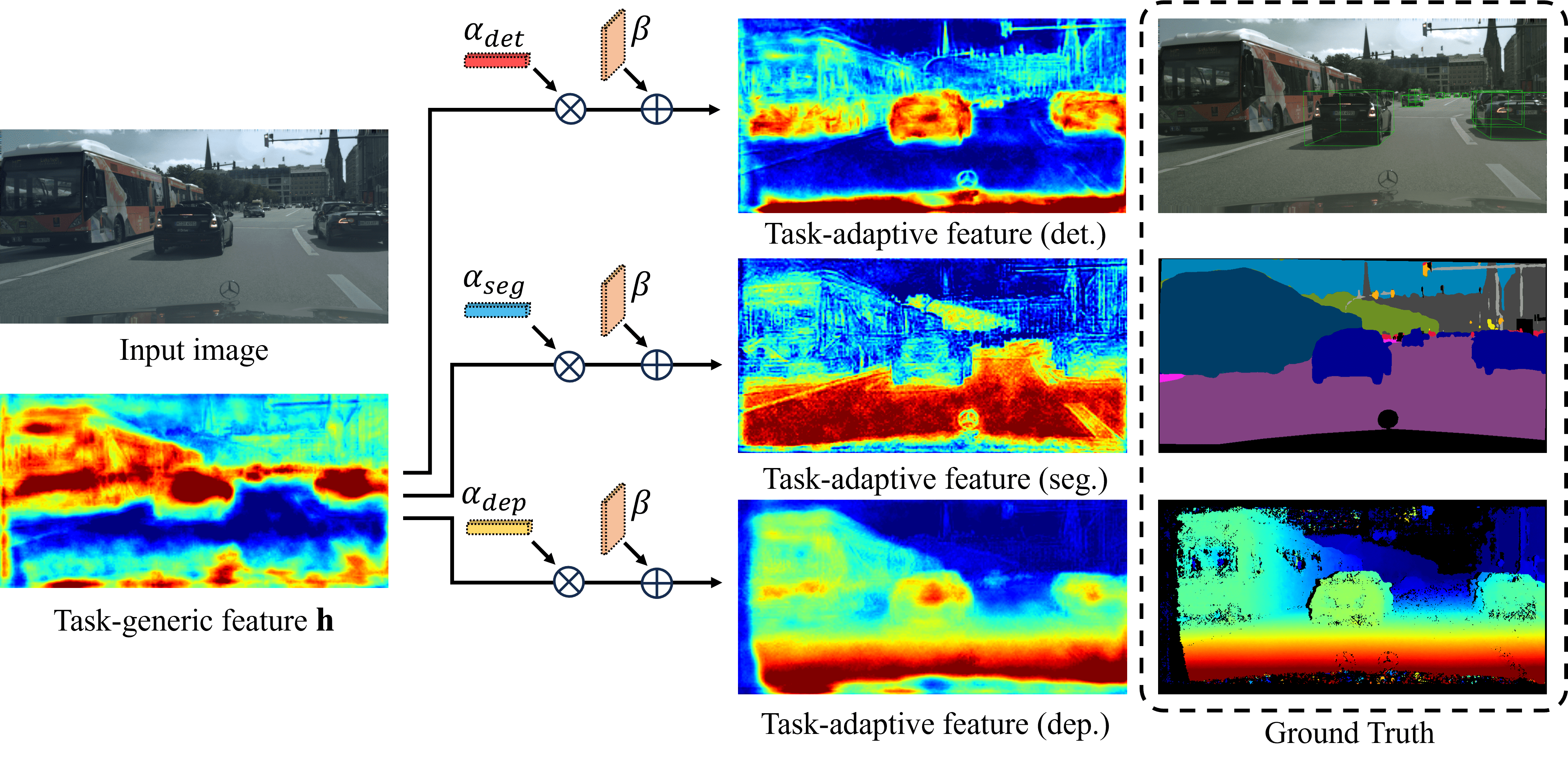}}
    \hfill
  \subfloat[Visualization of task-adaptive features w/wo the spatial attention.\label{1b}]{%
        \includegraphics[width=0.44\linewidth]{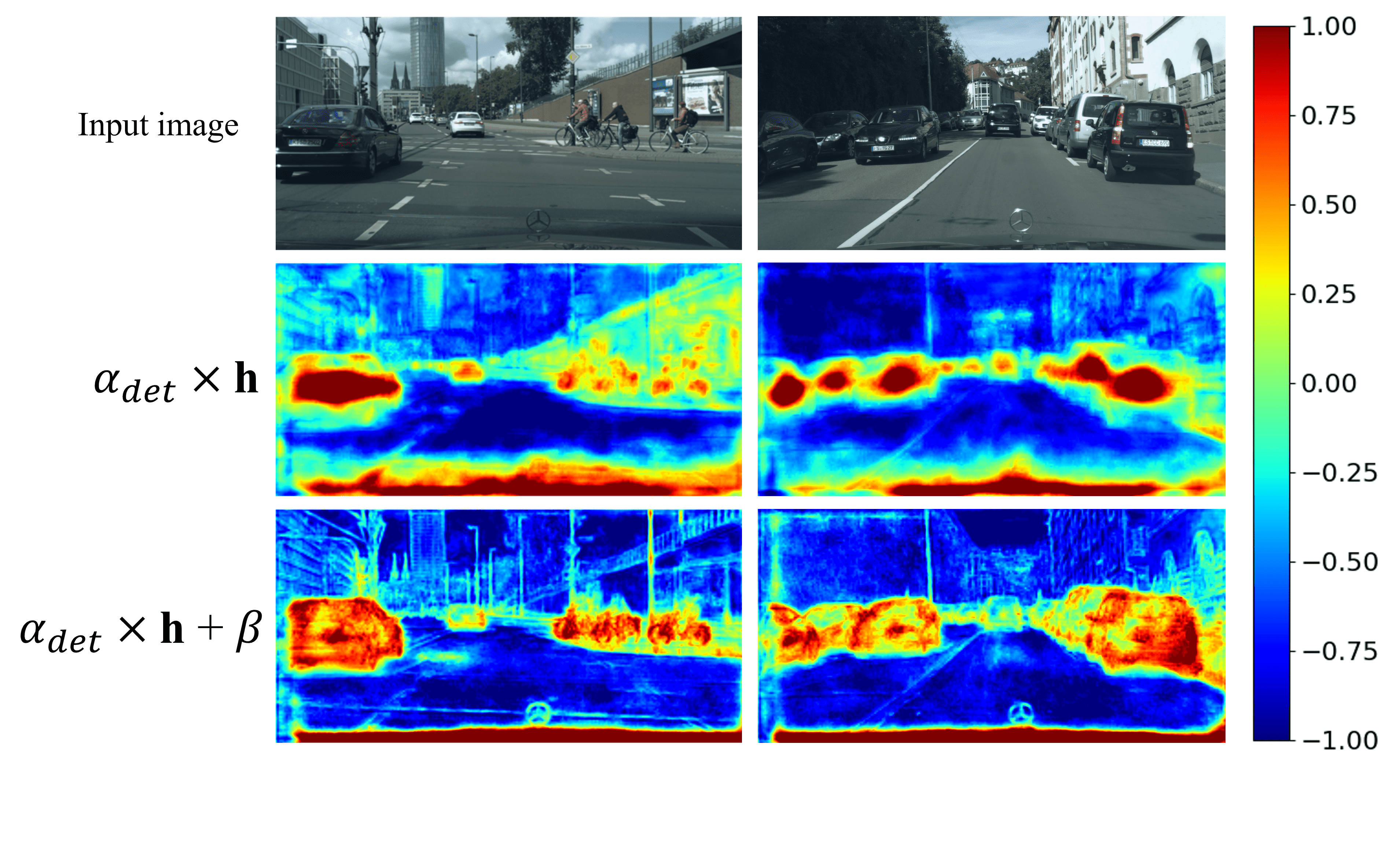}}
        \vspace{1pt}
  \caption{Feature heatmap visualization with respect to attention in TAG module.}
  \label{fig:attn} 
  \vspace{-5pt}
\end{figure*}

\subsection{Evaluation results on Cityscapes-3D}
To compare the MTL baselines and ours to single-task setups, we select the DLA34~\cite{yu2018deep} which is widely used in 3D object detection, as the benchmark for single-task setups. It is important to note that among all tasks, 3D object detection is the most computationally intensive.
To investigate the trade-off between performance and computational costs, we analyze the GFLOPs, number of parameters, and inference speed.
We measure the inference speed for all baselines on an NVIDIA Titan RTX GPU, with a batch size of 1.

Results detailed in \tabref{tab:Cityscapes-3D} highlight the prowess of the proposed architecture, especially in the detection task.
For semantic segmentation, the absence of the TAG module results in a performance drop of $-0.72\%$p and $-1.91\%$p when compared to the single-task configurations of DLA34 and DDRNet-39, respectively.
However, incorporating the TAG module elevates performance across all metrics, incurring only a slight rise in inference time by 0.56 ms. This increase is minimal when compared to the variant devoid of the TAG module.
Intriguingly, regardless of the presence of the TAG module, our method emerges as the sole configuration showcasing a surge in the averaged relative performance $\Delta \mathcal{T}$ relative to the single-task setup.
This amounts to a notable performance boost of $+7.4\%$p against the second-best model, DDRNet-39, while also achieving a 32.8\% reduction in inference time. 

\subsection{Additional experiments}

\subsubsection{Visualizing the impact of TAG modules}
To demonstrate the efficacy of channel attention within the TAG, we visualize task-adaptive features $\mathbf{h}_t$ produced from channel/spatial attention. These features are derived by using task-specific channel attention $\alpha_t$ and task-generic spatial attention $\beta$, with the task-generic feature $\mathbf{h}$.
Following this, we apply global channel pooling to condense the resultant features into a single channel for visualization purposes. As shown in \figref{fig:attn}-(a), it is evident that the TAG module can generate distinctive task-adaptive features.
For example, $\alpha_{det}$ mainly emphasizes prominent objects, highlighted in red on the feature map, and obstacles pertinent to the 3D detection task.
Conversely, $\alpha_{seg}$ focuses on the semantic regions, with a particular emphasis on the `road' class in the given example.
Meanwhile, $\alpha_{dep}$ adjusts the features to closely match the patterns observed in the depth ground truth.

Alongside our examination of the TAG module, we delve into the influence of task-generic spatial attention $\beta$ by contrasting the visualization of the feature map both with and without the incorporation of the attention.
As shown in \figref{fig:attn}-(b), the spatial attention drawing from detailed features of a high-resolution branch distinctly highlights spatial details. This is particularly prominent in object boundaries and smaller elements, notably in traffic participants such as cars and cyclists. Additionally, the intricate designs of poles and trees are faithfully retained. 
This underscores the effectiveness of spatial attention in directing the model to emphasize spatial data.

\subsubsection{Ablation studies on architectural elements}
To delve deeper into the individual contributions of specific architectural elements, we conduct comprehensive ablation studies.
Our design encapsulates four core components: the dual-branch structure, aggregation layer, semantic channel attention, and spatial attention.
In the experiment presented in \tabref{tab:ablation}, we either remove or replace one of these components for each trial (\textit{e.g.}, aggregation layer $\rightarrow$ Pyramid pooling module (PPM)~\cite{zhao2017pyramid}).
These experiments adopt the same configurations as those in \tabref{tab:Cityscapes-3D}.
The results consistently reveal that models lacking or altering one of our architectural components tend to underperform compared to our original proposal.
The proposed architecture with the TAG module achieves the best performance among the existing baselines.
This attests to the value and effectiveness of each subcomponent in enhancing performance within the MTL framework.

\begin{table}[t]
    \vspace{3pt}
    \caption{Ablation studies on architectural elements {\scriptsize(br: high-res branch, agg: aggregation layer, $\{\alpha_t, \beta\}$: channel/spatial attn).}}
    \label{tab:ablation}
    \centering
    \resizebox{0.99\linewidth}{!}{
    \begin{tabular}{ccccccccr}
    \toprule
    \multirow{2}{*}{br} & \multirow{2}{*}{agg} & \multirow{2}{*}{$\alpha_t$} & \multirow{2}{*}{$\beta$} & 3D Det. & Seg. & Dep. & \multirow{2}{*}{FPS} & \multirow{2}{*}{$\Delta_{\mathcal{T}}$ (\%)} \\
    \cmidrule{5-7}
    & & & & DS & mIoU & RMSE & & \\
    \midrule
    \multicolumn{4}{c}{DLA34 (STL)} & 54.55 & 66.38 & 5.50 & 17.4$\sim$17.8 & 0.00 \\
    \midrule
    & \checkmark & \checkmark & \checkmark & 49.13 & 58.99 & 6.21 & 37.1 & $-$11.33 \\
    \checkmark & & \checkmark & \checkmark & 49.10 & 60.09 & 5.74 & 35.0 & $-$7.94 \\
    \checkmark & \checkmark & & \checkmark & 55.12 & 61.02 & 5.56 & 30.3 & $-$2.71 \\
    \checkmark & \checkmark & \checkmark & & 58.21 & 63.71 & 5.62 & 30.7 & $+$0.17 \\
    \midrule
    \checkmark & \checkmark & \checkmark & \checkmark & \textbf{61.21} & \textbf{66.50} & \textbf{5.36} & 30.2 & $+$\textbf{4.98}\\
    \bottomrule
  \end{tabular}}
  \vspace{-7pt}
\end{table}

\vspace{5pt}
\section{Conclusion}

In this paper, we present a new multi-task learning approach tailored for real-time autonomous driving, encompassing monocular 3D object detection, semantic segmentation, and dense depth estimation.
Our uniquely designed network structure addresses the negative transfer problem inherent in heterogeneous multi-task learning, ensuring computational efficiency. Our proposed architecture, complemented by an attention-based module, capitalizes on shared knowledge across tasks, fostering task-adaptive learning. Drawing from domains of 3D object detection, semantic segmentation, and dense depth estimation, our architecture is optimized to adeptly handle per-pixel classification/regression tasks, accommodating objects of varied sizes. The TAG mechanism effectively accentuates task-relevant features, suppressing unrelated data. Our model's superior performance over baseline models highlights its pivotal role in advancing safe and efficient autonomous vehicles. We believe that this study signifies a substantial progression in multi-task learning tailored for real-time autonomous driving.

\vspace{1pt}
\section*{Acknowledgement}
This research was supported in part by Autonomous Driving Center, Hyundai Motor Company.
\clearpage
\bibliographystyle{IEEEtran}
\bibliography{IEEEabrv, egbib}

\end{document}